\documentclass[11pt,letterpaper]{article}
\usepackage{emnlp2017}
\usepackage{times}
\usepackage{latexsym}

\usepackage{caption}
\usepackage{amsmath}
\usepackage{algorithm}
\usepackage{algorithmic}
\usepackage{bm}
\usepackage{booktabs}
\usepackage{graphicx}
\def\emnlppaperid{361}

\emnlpfinalcopy

\title{Combining Generative and Discriminative Approaches to Unsupervised Dependency Parsing via Dual Decomposition\Thanks{This work was supported by the National Natural Science Foundation of China (61503248).}}

\author{Yong Jiang, Wenjuan Han \and Kewei Tu\\
    {\tt \{jiangyong,hanwj,tukw\}@shanghaitech.edu.cn}\\
  School of Information Science and Technology\\ShanghaiTech University, Shanghai, China\\}

\date{}

\begin{document}

\maketitle

\begin{abstract}
Unsupervised dependency parsing aims to learn a dependency parser from unannotated sentences. Existing work focuses on either learning generative models using the expectation-maximization algorithm and its variants, or learning discriminative models using the discriminative clustering algorithm. In this paper, we propose a new learning strategy that learns a generative model and a discriminative model jointly based on the dual decomposition method. Our method is simple and general, yet effective to capture the advantages of both models and improve their learning results. We tested our method on the UD treebank and achieved a state-of-the-art performance on thirty languages. 
\end{abstract}

\section{Introduction}
Dependency parsing is an important task in natural language processing. It identifies dependencies between words in a sentence, which have been shown to benefit other tasks such as semantic role labeling \cite{lei-EtAl:2015:NAACL-HLT} and sentence classification \cite{ma2015dependency}. Supervised learning of a dependency parser requires annotation of a training corpus by linguistic experts, which can be time and resource consuming. Unsupervised dependency parsing eliminates the need for dependency annotation by directly learning from unparsed text. 

Previous work on unsupervised dependency parsing mainly focuses on learning generative models, such as the dependency model with valence (DMV) \cite{klein2004corpus:1} and combinatory categorial grammars (CCG) \cite{bisk2012simple}. Generative models have many advantages. For example, the learning objective function can be defined as the marginal likelihood of the training data, which is typically easy to compute in a generative model. In addition, many types of inductive bias, such as those favoring short dependency arcs \cite{smith2006annealing}, encouraging correlations between POS tags \cite{cohen2008logistic,cohen2009shared,berg2010painless,jiang-han-tu:2016:EMNLP2016}, and limiting center embedding \cite{noji-miyao-johnson:2016:EMNLP2016}, can be incorporated into generative models to achieve better parsing accuracy. However, due to the strong independence assumption in most generative models, it is difficult for these models to utilize context information that has been shown to benefit supervised parsing.

Recently, a feature-rich discriminative model for unsupervised parsing is proposed that captures the global context information of sentences \cite{grave-elhadad:2015:ACL-IJCNLP}. Inspired by discriminative clustering, learning of the model is formulated as convex optimization of both the model parameters and the parses of training sentences. By utilizing language-independent rules between pairs of POS tags to guide learning, the model achieves state-of-the-art performance on the UD treebank dataset. 

In this paper we propose  to jointly train two state-of-the-art models of unsupervised dependency parsing: a generative model called LC-DMV \cite{noji-miyao-johnson:2016:EMNLP2016} and a discriminative model called Convex-MST \cite{grave-elhadad:2015:ACL-IJCNLP}. We employ a learning algorithm based on the dual decomposition \cite{dantzig1960decomposition} inference algorithm, which encourages the two models to influence each other during training.

We evaluated our method on thirty languages and found that the jointly trained models surpass their separately trained counterparts in parsing accuracy. Further analysis shows that the two models positively influence each other during joint training by implicitly sharing the inductive bias. 

\section{Preliminaries}

\subsection{DMV}
The dependency model with valence (DMV) \cite{klein2004corpus:1} is the first generative model that outperforms the left-branching baseline in unsupervised dependency parsing. 
In DMV, a sentence is generated by recursively applying three types of grammar rules to construct a parse tree from the top down.
The probability of the generated sentence and parse tree is the probability product of all the rules used in the generation process. 
To learn the parameters (rule probabilities) of DMV, the expectation maximization algorithm is often used.
Noji et al. \shortcite{noji-miyao-johnson:2016:EMNLP2016} exploited two universal syntactic biases in learning DMV: restricting the center-embedding depth and encouraging short dependencies.
They achieved a comparable performance with state-of-the-art approaches. 

\subsection{Convex-MST}
\label{sec:convex_mst}
Convex-MST \cite{grave-elhadad:2015:ACL-IJCNLP} is a discriminative model for unsupervised dependency parsing based on the first-order maximum spanning tree dependency parser \cite{mcdonald2005online}. 
Given a sentence, whether each possible dependency exists or not is predicted based on a set of handcrafted features and a valid parse tree closest to the prediction is identified by the minimum spanning tree algorithm.

For each sentence $\mathbf{x}$, a first-order dependency graph is built over the words of the sentence. The weight of each edge is calculated by $\mathbf{w}^{T}\mathbf{f}(\mathbf{x},i,j)$, where $\mathbf{w}$ is the parameters and $\mathbf{f}(\mathbf{x},i,j)$ is the handcrafted feature vector of the dependency from the $i$-th word to the $j$-th word in sentence $\mathbf{x}$. For sentence $\mathbf{x}$ of length $n$, we can represent it as matrix $\mathbf{X}$ where each raw is a feature vector. The parse tree $\mathbf{y}$ is a spanning tree of the graph and can be represented as a binary vector with length $n\times n$ where each element is 1 if the corresponding arc is in the tree and 0 otherwise. 

Learning is based on discriminative clustering with the following objective function: 
 \[
  \frac{1}{N} \sum_{\alpha = 1}^{N} \left( \frac{1}{2n_{\alpha}} ||\mathbf{y}_{\alpha} - \mathbf{X}_{\alpha}\mathbf{w} ||_2^2 - \mu \mathbf{v}^{T}\mathbf{y}_{\alpha}\right) + \frac{\lambda}{2} ||\mathbf{w}||_2^2 
 \]
where $\mathbf{X}_{\alpha}$ is a matrix where each row is a feature representation $\mathbf{f}(\mathbf{x}_{\alpha},i,j)$ of an edge in the dependency graph of sentence $\mathbf{x}_{\alpha}$, $\mathbf{v}$ represents whether each dependency arc in $\mathbf{y}_{\alpha}$ satisfies a set of pre-specified linguistic rules, and $\lambda$ and $\mu$ are hyper-parameters. The Frank-Wolfe algorithm is employed to optimize the objective function.

\subsection{Dual Decomposition}
Dual decomposition \cite{dantzig1960decomposition}, a special case of Lagrangian relaxation, is an optimization method that decomposes a hard problem into several small sub-problems. It has been widely used in machine learning \cite{komodakis2007mrf} and natural language processing \cite{koo2010dual,rush2012tutorial}. 

Komodakis et al. \shortcite{komodakis2007mrf} proposed using dual decomposition to do MAP inference for Markov random fields. Koo et al. \shortcite{koo2010dual} proposed a new dependency parser based on dual decomposition by combining a graph based dependency model and a non-projective head automata. In the work of Rush et al. \shortcite{rush2010dual}, they showed that dual decomposition can effectively integrate two lexicalized parsing models or two correlated tasks.

\subsection{Agreement based Learning}
Liang et al. \shortcite{NIPS2007_3246} proposed agreement based learning that trains several tractable generative models jointly and encourages them to agree on certain latent variables. To effectively train the system, a product EM algorithm was used. They showed that the joint model can perform better than each independent model on the accuracy or convergence speed.  They also showed that the objective function of the work of Klein and Manning \shortcite{klein2004corpus:1} is a special case of the product EM algorithm for grammar induction. 
Our approach has a similar motivation to agreement based learning but has two important advantages. First, while their approach only combines generative models, our approach can make use of both generative and discriminative models. Second, while their approach requires the sub-models to share the same dynamic programming structure when performing decoding, our approach does not have such restriction. 

\section{Joint Training}
We minimize the following objective function that combines two different models of unsupervised dependency parsing:
\begin{align*}
& J(\mathbf{M_F}, \mathbf{M_G}) \\
& = \sum_{\alpha = 1}^{N} \min_{\mathbf{y_{\alpha} \in \mathcal{Y}_\alpha} } \left( F(\mathbf{x_{\alpha}}, \mathbf{y_{\alpha}}; \mathbf{M_F}) + G(\mathbf{x_{\alpha}}, \mathbf{y_{\alpha}}; \mathbf{M_G}) \right)
\end{align*}
where $N$ is the size of training data, $\mathbf{M_F}$ and $\mathbf{M_G}$ are the parameters of the first and second model respectively, $F$ and $G$ are their respective learning objectives, and $\mathcal{Y}_\alpha$ is the set of valid dependency parses of sentence $\mathbf{x_{\alpha}}$.
While in principle this objective can be used to combine many different types of models, here we consider two state-of-the-art models of unsupervised dependency parsing, a generative model LC-DMV \cite{noji-miyao-johnson:2016:EMNLP2016} and a discriminative model Convex-MST \cite{grave-elhadad:2015:ACL-IJCNLP}. We denote the parameters of LC-DMV by $\Theta$ and the parameters of Convex-MST by $\mathbf{w}$. Their respective objective functions are,
\begin{align*}
& F(\mathbf{x_{\alpha}}, \mathbf{y_{\alpha}}; \Theta) = - \log \left(P_{\Theta} (\mathbf{x}_{\alpha}, \mathbf{y}_{\alpha}) f(\mathbf{x}_{\alpha}, \mathbf{y}_{\alpha})\right) \\
& G(\mathbf{x_{\alpha}}, \mathbf{y_{\alpha}}; \mathbf{w}) \\
& = \frac{1}{2n_{\alpha}} ||\mathbf{y}_{\alpha} - \mathbf{X}_{\alpha} \mathbf{w}||_2^2 + \frac{\lambda}{2N} ||\mathbf{w}||_2^2 -\mu \mathbf{v}^T\mathbf{y}
\end{align*}
where $P_{\Theta}(\mathbf{x}_{\alpha}, \mathbf{y}_{\alpha})$ is the joint probability of sentence $\mathbf{x_{\alpha}}$ and parse $\mathbf{y_{\alpha}}$, $f$ is a constraint factor, and the notations in the second objective function are explained in section \ref{sec:convex_mst}.

\subsection{Learning}
We use coordinate descent to optimize the parameters of the two models. In each iteration, we first fix the parameters and find the best dependency parses of the training sentences (see section \ref{sec:decode}); we then fix the parses and optimize the parameters.
The detailed algorithm is shown in Algorithm \ref{algo:param_learn}.

Pretraining of the two models is done by running their original learning algorithms separately. When the parses of the training sentences are fixed, it is easy to show that the parameters of the two models can be optimized separately. Updating the parameters $\Theta$ of LC-DMV can be done by simply counting the number of times each rule is used in the parse trees and then normalizing the counts to get the maximum-likelihood probabilities. The parameters $\mathbf{w}$ of Convex-MST can be updated by stochastic gradient descent. After updating $\Theta$ and $\mathbf{w}$ at each iteration, we additionally train each model separately for three iterations, which we find further improves learning. 

\begin{algorithm}[t]
	\caption{Parameter Learning}
	\label{algo:param_learn}
	\begin{algorithmic}
		\STATE {\bfseries Input:} Training sentence $\mathbf{x}_1, \mathbf{x}_2, ..., \mathbf{x}_N$
        \STATE Pre-train $\Theta$ and $\mathbf{w}$
		\REPEAT
        \STATE Fix $\Theta$ and $\mathbf{w}$ and solve the decoding problem $~~~~~~$ to get $\mathbf{y}_{\alpha}, \alpha = 1, 2, \ldots, N$
		\STATE Fix the parses and update $\Theta$ and $\mathbf{w}$
		\UNTIL{Convergence}
	\end{algorithmic}
\end{algorithm}

\subsection{Joint Decoding}\label{sec:decode}
Given a training sample $\mathbf{x}$ and parameters $\mathbf{w},\Theta$, the goal of decoding is to find the best parse tree:
\[\hat{\mathbf{y}} = \arg\min_{\mathbf{y} \in \mathcal{Y}} \frac{1}{2n} ||\mathbf{y} - \mathbf{X}\mathbf{w}||_2^2 - \mu \mathbf{v}^T \mathbf{y} - \log P_{\Theta} (\mathbf{x}, \mathbf{y})\]
We employ the dual decomposition algorithm to solve this problem (shown in Algorithm \ref{algo:DD}), where $\tau$ represents the step size.
\begin{algorithm}[t]
	\caption{Decoding via Dual Decomposition}
	\label{algo:DD}
	\begin{algorithmic}
		\STATE {\bfseries Input:} Sentence $\mathbf{x}$, fixed parameters $\mathbf{w}$ and $\Theta$
        \STATE Initialize vector $\mathbf{u}$ of size $n\times n$ to $\mathbf{0}$
		\REPEAT
       \STATE $\hat{\mathbf{y}} = \arg\min_{\mathbf{y} \in \mathcal{Y}} F(\mathbf{x}, \mathbf{y}; \Theta) + \mathbf{u}^{T}\mathbf{y}$
        \STATE $\hat{\mathbf{z}} = \arg\min_{\mathbf{z} \in \mathcal{Y}} G(\mathbf{x}, \mathbf{z}; \mathbf{w}) - \mathbf{u}^{T} \mathbf{z}$
		\IF{$\hat{\mathbf{y}} = \hat{\mathbf{z}}$}
        	\STATE return $\hat{\mathbf{y}}$
        \ELSE
			\STATE $\mathbf{u}=\mathbf{u} + \tau \left(\hat{\mathbf{y}}-\hat{\mathbf{z}}\right)$
        \ENDIF
		\UNTIL{Convergence}
	\end{algorithmic}
\end{algorithm}

The most important part of the algorithm is solving the two separate decoding problems:
\[
\hat{\mathbf{y}} = \arg\min_{\mathbf{y}\in \mathcal{Y}} - \log (P_{\Theta} (\mathbf{x}, \mathbf{y}) f(\mathbf{x}, \mathbf{y})) + \mathbf{u}^{T} \mathbf{y}\\
\]
\[
\hat{\mathbf{z}} = \arg\min_{\mathbf{z} \in \mathcal{Y}} \frac{1}{2n} ||\mathbf{z} - \mathbf{X}\mathbf{w}||_2^2 - \mu \mathbf{v}^T \mathbf{z} - \mathbf{u}^{T} \mathbf{z}
\]
The first decoding problem can be solved by a modified CYK parsing algorithm that takes into account the information in vector $\mathbf{u}$. The second decoding problem can be solved using the same algorithm of Grave and Elhadad \shortcite{grave-elhadad:2015:ACL-IJCNLP} (we use the projective version in our approach).

\section{Experiments}
\subsection{Setup}
We use UD Treebank 1.4 as our datasets. We sorted the datasets in the treebank by the number of training sentences of length $\leq$ 15 and selected the top thirty datasets, which is similar to the setup of Noji et al. \shortcite{noji-miyao-johnson:2016:EMNLP2016}. For each dataset, we trained our method on the training data with length $\leq$ 15 and tested our method on the testing data with length $\leq$ 40. We tuned the hyper-parameters of our method on the dataset of the English language and reported the results on the thirty datasets without any further parameter tuning. We compared our method with four baselines. The first two baselines are Convex-MST and LC-DMV that are independently trained. 
To construct the third baseline, we used the independently trained Convex-MST baseline to parse all the training sentences and then used the parses to initialize the training of LC-DMV. This can be seen as a simple method to combine two different approaches.
On the other hand, we did not use the LC-DMV baseline to initialize Convex-MST training because the objective function of Convex-MST is convex and therefore the initialization does not matter.

\subsection{Results}
In Table \ref{tab:mul_ling}, we compare our jointly trained models with the four baselines. We can see that with joint training and independent decoding, LC-DMV and Convex-MST can achieve superior overall performance than when they are separately trained with or without mutual initialization. 
 Joint decoding with our jointly trained models performs worse than independent decoding. We made the same observation when applying joint decoding to the separately trained models (not shown in the table). 
We believe this is because unsupervised parsers have relatively low accuracy and forcing them to reconcile would not lead to better parses. On the other hand, joint decoding during training helps propagate useful inductive biases between models and thus leads to better trained models.

\begin{table}[t]
\setlength\tabcolsep{4pt}
\small
\captionsetup{font=small}
\centering
\begin{tabular}{l||c|c||c||c|c|c}
\hline
Language & M & D & D-I & M-J & D-J & DD \\ \hline
A\_Greek & 43.4 & 33.1 & 38.8 & 44.2 & \bf 44.9 & 38.9 \\ \hline
A\_Greek-P & 50.4 & 43.0 & 44.7 & 50.8 & \bf 52.9 & 44.9 \\ \hline
Basque & 50.0 & 45.4 & 54.2 & 52.1 & \bf 55.7 & 50.2 \\ \hline
Bulgarian & 61.6 & 62.4 & 60.3 & 64.7 & \bf 73.8 & 64.8 \\ \hline
Czech & 48.6 & 17.4 & 53.9 & 48.7 & \bf 54.0 & 53.5 \\ \hline
Czech-CAC & 50.4 & 53.0 & 53.9 & 55.6 & \bf 62.3 & 50.2 \\ \hline
Dutch & 45.3 & 34.1 & \bf 56.7 & 48.2 & 43.5 & 40.7 \\ \hline
Dutch-LS & 42.4 & 27.0 & 16.4 & \bf 43.2 & 41.2 & 36.3 \\ \hline
English & 54.0 & 56.0 & 49.8 & 57.3 & \bf 60.1 & 53.4 \\ \hline
Estonian &\bf  49.4 & 31.8 & 47.5 & 48.7 & 44.0 & 44.4 \\ \hline
Finnish &\bf 44.7 & 26.9 & 39.0 & 44.2 & 43.5 & 31.2 \\ \hline
Finnish-FTB &\bf 49.9 & 31.0 & 47.9 & 47.7 & 48.0 & 36.5 \\ \hline
French &\bf 62.0 & 48.6 & 57.0 & 54.5 & 57.0 & 55.5 \\ \hline
German & 51.4 & 50.5 & 54.1 & 49.3 & \bf 55.7 & 48.6 \\ \hline
Gothic & 52.7 & 49.9 & 47.3 & \bf 59.6 & 56.4 & 58.0 \\ \hline
Hindi & 56.8 & 54.2 & 48.4 & 52.1 & \bf 60.0 & 49.1 \\ \hline
Italian & 69.1 & \bf 71.1 & 67.4 & 62.8 & 70.3 & 64.5 \\ \hline
Japanese & 44.8 & 43.8 & 43.8 & 42.8 & \bf 45.8 & 41.0 \\ \hline
Latin-ITTB & 38.8 & 38.6 & 42.3 & \bf 47.0 & 42.2 & 40.3 \\ \hline
Latin-PROIEL & 44.3 & 34.8 & 38.7 & \bf 46.8 & 41.8 & 42.9 \\ \hline
Norwegian & 55.3 & 45.5 & 51.4 & 57.4 & \bf 60.8 & 46.6 \\ \hline
Old\_Church\_S & 56.4 & 26.6 & 51.3 & 58.3 & \bf 58.6 & 42.0 \\ \hline
Polish & 63.4 & 63.7 & 61.5 & 70.7 & \bf 74.2 & 68.9 \\ \hline
Portuguese & 57.9 & \bf 67.2 & 60.1 & 56.1 &  62.9 & 57.4 \\ \hline
Portuguese-BR & 59.3 & 63.1 & 62.0 & 65.5 & \bf 68.8 & 58.3 \\ \hline
Russian-STR & 47.6 & 51.7 & 56.5 & 52.1 & \bf 64.4 & 52.6 \\ \hline
Slovak & 57.4 & 59.3 & 51.9 & 61.7 & \bf 65.9 & 58.7 \\ \hline
Slovenian & 54.0 & 49.5 & 56.3 & 65.5 & \bf 69.6 & 56.1 \\ \hline
Spanish & 61.9 & 61.9 & 60.3 & 57.4 & \bf 68.0 & 60.2 \\ \hline
Spanish-AC & 59.4 & 59.5 & 56.4 & 56.8 & \bf 65.2 & 57.6 \\ \hline
\hline
Average & 52.7 & 47.2 & 50.3 & 54.2 & \bf 56.5 & 49.6 \\ \hline
\hline
Average $\leq 15$ & 55.4 & 48.9 & 54.9 & 57.3 & \bf 60.2 & 53.8 \\ \hline
\end{tabular}
\caption{Directed dependency accuracy on thirty datasets with test sentences of length $\leq$ 40. The last row indicates the average directed accuracy on sentences of length $\leq$ 15. M (Convex-MST) and D (LC-DMV) are the independently trained baselines. D-I is the third baseline in which the LC-DMV training is initialized by the parses produced from the trained Convex-MST model. With our jointly trained models, M-J and D-J denote separate decoding and DD denotes joint decoding.}
\label{tab:mul_ling}
\end{table}

\subsection{Analysis of Parsing Results}
We analyze the parsing results from the two models to see how they benefit each other with joint training. Note that LC-DMV limits the depth of center embedding and encourages shorter dependency length, while Convex-MST encourages dependencies satisfying pre-specified linguistic rules. Therefore, we would like to see whether the jointly-trained LC-DMV produces more dependencies satisfying the linguistic priors than its separately-trained counterpart, and whether the jointly-trained Convex-MST produces parse trees with less center embedding and shorter dependencies than its separately-trained counterpart.

\begin{figure}[t]
	\centering
    \captionsetup{font=small}
	\includegraphics[scale=0.4]{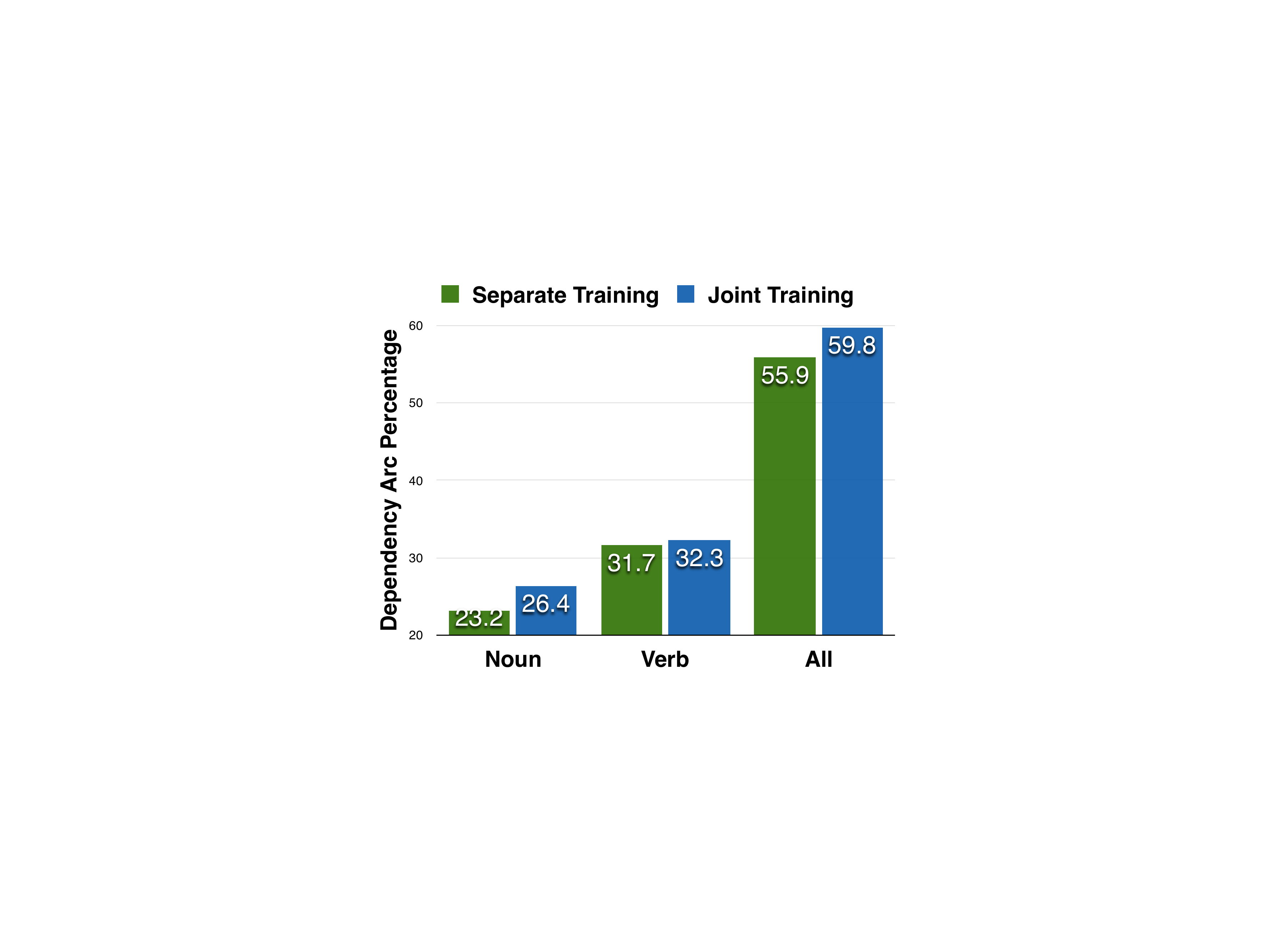}
	\caption{Percentages of dependencies satisfying linguistic rules in the LC-DMV parses of the English test dataset. Noun and Verb denote dependencies headed by nouns and verbs.}
	\label{fig:rule}
\end{figure}

Figure \ref{fig:rule} shows the percentages of dependencies satisfying linguistic rules when using the separately and jointly trained LC-DMV to parse the test sentences in the English dataset. As we can see, with joint training, LC-DMV is indeed influenced by Convex-MST and produces more dependencies satisfying linguistic rules.

Table \ref{tab:arc} shows the average dependency length when using the separately and jointly trained Convex-MST to parse the English test dataset. The dependency length can be seen to decrease with joint training, showing the influence from LC-DMV. As to center embedding depth, we find that separately trained Convext-MST already produces very few center embeddings of depth 2 or more, so the influence from the center embedding constraint of LC-DMV during joint training is not obvious.
We note that the influence on Convex-MST from LC-DMV during joint training is relatively small, which may contribute to the much smaller accuracy improvement (1.5\%) of Convex-MST with joint training in comparison with the 9.3\% improvement of LC-DMV.
We conducted an additional experiment that scaled down the Convex-MST objective in joint training in order to increase the influence of LC-DMV. The results show that LC-DMV indeed influences Convex-MST to a greater degree, but the parsing accuracies of the two models decrease.


\begin{table}[t]
    	\centering
        \captionsetup{font=small}
    	\small
    	\begin{tabular}{|c|c|}
    		\hline
    		{\bf Methods} & {\bf  Average Dependency Length}\\\hline
    		Separate Training & 1.673\\
    		Joint Training & \bf 1.627 \\\hline 
    	\end{tabular}
      	\caption{Average dependency length in the Convex-MST parses of the English test dataset.}
      	\label{tab:arc}
    \end{table}

\section{Conclusion}
In this paper, we proposed a new learning strategy for unsupervised dependency parsing that learns a generative model and a discriminative model jointly based on dual decomposition. We show that with joint training, two state-of-the-art models can positively influence each other and achieve better performance than their separately trained counterparts. 


\bibliography{emnlp2017}
\bibliographystyle{emnlp_natbib}

\end{document}